\documentclass[10pt,twocolumn,letterpaper]{article}

\usepackage{algorithm}
\usepackage{algpseudocode}
\usepackage{multirow}

\usepackage[pagenumbers]{cvpr} % To force page numbers, e.g. for an arXiv version

\definecolor{cvprblue}{rgb}{0.21,0.49,0.74}
\usepackage[pagebackref,breaklinks,colorlinks,allcolors=cvprblue]{hyperref}

\def\paperID{16773} % *** Enter the Paper ID here
\def\confName{CVPR}
\def\confYear{2025}

\title{BioX-CPath: Biologically-driven Explainable Diagnostics for Multistain IHC Computational Pathology}

\author{
Amaya Gallagher-Syed\thanks{Queen Mary University of London} \thanks{Corresponding author: a.r.syed@qmul.ac.uk} \hspace{1em} 
Henry Senior\textsuperscript{*} \hspace{1em}
Omnia Alwazzan\textsuperscript{*} \hspace{1em}
Elena Pontarini\textsuperscript{*}\\[1ex]
Michele Bombardieri\textsuperscript{*} \hspace{1em}
Costantino Pitzalis\textsuperscript{*} \hspace{1em}
Myles J. Lewis\textsuperscript{*} \hspace{1em}
Michael R. Barnes\textsuperscript{*}\\[1ex]
\hspace{2em}Luca Rossi\thanks{The Hong Kong Polytechnic University} \hspace{4em}
Gregory Slabaugh\textsuperscript{*}\\[2ex]
}
\begin{document}
\maketitle
\begin{abstract}
The development of biologically interpretable and explainable models remains a key challenge in computational pathology, particularly for multistain immunohistochemistry (IHC) analysis. We present BioX-CPath, an explainable graph neural network architecture for whole slide image (WSI) classification that leverages both spatial and semantic features across multiple stains. At its core, BioX-CPath introduces a novel Stain-Aware Attention Pooling (SAAP) module that generates biologically meaningful, stain-aware patient embeddings. Our approach achieves state-of-the-art performance on both Rheumatoid Arthritis and Sjogren's Disease multistain datasets. Beyond performance metrics, BioX-CPath provides interpretable insights through stain attention scores, entropy measures, and stain interaction scores, that permit measuring model alignment with known pathological mechanisms. This biological grounding, combined with strong classification performance, makes BioX-CPath particularly suitable for clinical applications where interpretability is key. Source code and documentation can be found at: \url{https://github.com/AmayaGS/BioX-CPath}\footnotetext{Accepted for publication at CVPR 2025}.
\end{abstract}
    
\section{Introduction}
\label{sec:intro}

Whole Slide Image (WSI) scanners capture high resolution, multi-magnification digital images of stained tissue biopsies presented on glass slides. The digitization of these biopsies has spurred the development of computational pathology methods. Analysis of these WSIs currently stands as one of the gold standard diagnostic and subtyping methods for many forms of cancers and autoimmune diseases, such as Rheumatoid Arthritis (RA) and Sjogren's Disease. Different types of staining exist, which highlight different aspects of the tissue samples. Hematoxylin \& Eosin (H\&E) staining, a traditional and widely used technique, offers a broad view of tissue architecture and cellular morphology, with Hematoxylin staining cell nuclei a deep blue-purple, while Eosin stains cytoplasm and extracellular matrix in shades of pink. In contrast, Immunohistochemistry (IHC) is a more specialized technique that uses antibodies tagged with visual markers to identify specific proteins or cell types within tissue samples, allowing for precise localization and visualization of cell populations present in the tissue \cite{Magaki2019}. 

In cancer diagnostics, H\&E staining remains the foundation for initial assessment and general diagnosis. However, IHC plays a crucial role in tumor classification, prognosis determination, and treatment selection by pinpointing specific cancer biomarkers, such Human Epidermal Growth Factor Receptor 2 (HER2), Estrogen Receptor (ER) and Progesterone Receptor (PR) \cite{Zhou2024,ZilenaitePetrulaitiene2024}. For autoimmune diseases, while H\&E staining identifies general patterns of inflammation and tissue damage, IHC becomes essential for a more nuanced understanding of the disease process. It highlights the types of immune cells present in inflammatory infiltrates, detects autoantibody deposits, and visualizes specific autoantigens targeted by the immune system \cite{Pitzalis2014}. In clinical pathology, a tissue sample will be taken and thinly sliced, and different stains applied to these slices, often with a reference H\&E slide to verify tissue quality \cite{Magaki2019}. These multi-stain WSI stacks are rich in information about cellular types, tissue structures, and spatial patterns which relate to disease presentation and prognosis. Expert pathologists examining these stacks perform a semi-quantitative analysis, efficiently integrating information across both scale, stains, and images.

Most state-of-the-art computational pathology methods so far have focused on H\&E and the single-stain domain. Work that has tackled IHC has often done so in the context of cell quantification via cell segmentation \cite{ihc_nuclei_1,ihc_nuclei_2,ihc_nuclei_3,madeleine}, as well as prediction and scoring of biomarkers \cite{biomarkers_1, Kather2019,madeleine}, or registration of multistain stacks \cite{registration}. Some methods have also explored the potential of H\&E to IHC virtual staining techniques \cite{Spatial_lymp,virtual_ihc,virtual_ihc_2} or recently of using IHC as views for self-supervised representation learning \cite{madeleine}. Most of these approaches have concentrated on extracting information from IHC slides such that this could be predicted using H\&E or on quantification of cell populations in IHC. This is because H\&E is an older, more widely available and cost-effective technology. However, the use of IHC and more advanced techniques such as immunofluorescence is only set to grow in the coming years, associated with a decrease in technology cost and more advanced biomarker detection techniques \cite{Magaki2019}. There is therefore a clear need for methods which explicitly focus on integrating the complex cell landscapes across stains. To the best of our knowledge, few studies have concentrated on the issue of classification of unregistered, unannotated multistain datasets to date, with a single stain graph and mid/late fusion approaches developed in \cite{Dwivedi2022} and a multistain attention graph approach proposed in MUSTANG \cite{AGS23_BMVC}. However, the value of computational pathology extends beyond classification tasks. Given the rich information contained in multi-stain data, these methods must be interpretable and generate actionable biological insights at both the disease and patient level. This interpretability serves two crucial purposes: it advances our understanding of underlying disease mechanisms, and it allows pathologists to verify that model predictions align with established biological knowledge, building trust in the system's outputs.

\subsection{Contributions}

\begin{enumerate}

    \item We introduce BioX-CPath, a biologically-driven graph-based model tailored to the complex cellular landscapes in multistain datasets. BioX-CPath works across multiple stains using semantic and spatial cues to capture complementary cellular and tissue information.

    \item We propose a novel Stain-Aware Attention Pooling (SAAP) module that generates expressive, stain-aware patient embeddings. This module uniquely respects the biological and diagnostic diversity across stains, improving interpretability and diagnostic relevance.
    
    \item We fully leverage the biological interpretability of BioX-CPath via derived metrics: stain attention and entropy scores, stain-stain interaction scores and Graph Neural Networks (GNNs) node heatmaps. These metrics provide detailed insights into stain relevance and inter-stain relationships, uncovering key biological patterns and interactions that contribute to disease pathology. 

\end{enumerate}

\subsection{Related Work}

\subsubsection{Multiple Instance Learning}

WSIs are gigapixel, heterogeneous image files, which present challenges for computer vision methods given each image can reach over $100k\times 100k$ pixels, generally within a low patient sample setting. A weakly supervised Multiple Instance Learning (MIL) approach is most often employed to address this challenge. The image is divided into a regular grid of smaller patches (e.g., $224 \times 224$ pixels), each inheriting slide/patient labels. Patches are then embedded into a feature vector and classified at the slide/patient level using some form of non-trainable (e.g., max or mean) or trainable aggregation on the set of instances. Methods such as ABMIL \cite{Ilse18}, DS-MIL \cite{Li2021}, and CLAM introduced trainable linear attention aggregation layers \cite{Lu2021}. TransMIL \cite{shao2021transmil} tackles the issue of long-range dependencies by approximating self-attention operations between patches via the Nyström method \cite{Xiong2021}.

\subsubsection{GNNs in Histopathology}

Applications in histopathology can be divided into cell, patch, or tissue-level graphs, with both node and graph classification approaches being employed. Patch-graphs can be constructed using features extracted from a WSI or a set of WSIs, then connected via edges \cite{Achanta2012,Adnan2020,Zheng2022,Zheng2022a}. DeepGraphConv \cite{Li2018}, PatchGCN \cite{Chen2021a}, GTP \cite{Zheng2022}, CAMIL \cite{Fourkioti2023} and HEAT \cite{heat} adopt this approach by constructing a graph connecting either the $k$-nearest neighbors in feature space or region adjacent patches. DeepGraphConv uses spectral graph convolution on a subset of patches, whereas Patch-GCN employs graph convolutional layers with residual connections and a final global attention pooling mechanism layer, which GTP replaces with a Transformer layer. CAMIL \cite{Fourkioti2023} combines a spatial neighbor constrained attention module with a transformer layer. HEAT \cite{heat} incorporates node and edge attributes in a heterogeneous graph, together with a pseudo-label pooling algorithm based on predicted cell types using KimiaNet, a feature extractor which was pretrained on H\&E images from The Cancer Genome Atlas Program (TCGA) \cite{kimia}.  

These methods were designed for accurate classification and prognosis in H\&E staining and cancer datasets. Because of this, these models are optimized to focus on features and patterns linked to tissue architecture and cell morphology. Notably, because these models were designed for the single-stain cancer domain, they concentrate on spatial awareness which aligns well with the need in cancer to accurately detect tumors and tumor microenvironment based on tissue architecture and cell morphology \cite{Chen2021a,Lee2022, heat}. Moreover, these approaches provide insight into tumor localization by providing heatmaps overlays showing the attention scores obtained per patch. Other methods, such as TEA-Graph \cite{Lee2022} and Slide-Graph \cite{slide_graph} also provide insight into interpretable prognostic biomarkers linked to tissue type. 

In line with previous work, we adopt a patch-graph approach to efficiently integrate information across multistain WSI stacks. However, although we provide insight into the model decision making process through examination of layer importance and GNN heatmaps, our focus is on providing insight into the alignment of our model with underlying biology and in understanding how the cell populations interact. This approach bridges the gap between performance-based approaches and explainable, insight-driven approaches.
\section{Preliminaries}
\label{sec:preliminaries}

In this section we provide definitions and background on the concepts used throughout this work. 

\paragraph{Graph Neural Networks.} GNNs are capable of learning representations of graphs by propagating node features through a series of computationally efficient message-passing and aggregation operations \cite{Dwivedi2022a}. Given a graph over a set of nodes $V$, during the $k$-th message-passing iteration, the embedding $\mathbf{h}_u^{(k)}$ corresponding to each node $u \in V$ is updated according to information aggregated from the neighbors of $u$, i.e.,
\begin{equation}
\begin{aligned}
\mathbf{h}_u^{(k+1)} & =\operatorname{UPDATE}^{(k)}\left(\mathbf{h}_u^{(k)}, \mathbf{m}_{\mathcal{N}(u)}^{(k)}\right) \\
\mathbf{m}_{\mathcal{N}(u)}^{(k)} & = \operatorname{AGGREGATE}^{(k)}\left(\left\{\mathbf{h}_v^{(k)}, \forall v \in \mathcal{N}(u)\right\}\right)\, , 
\end{aligned}
\end{equation}
where the neighborhood $\mathcal{N}(u)$ is defined as the set of nodes that share an edge with $u$, UPDATE and AGGREGATE are arbitrary differentiable functions, and $\mathbf{m}_{\mathcal{N}(u)}^{(k)}$ is the ``message'' that is aggregated from $\mathcal{N}(u)$. At each iteration, the AGGREGATE function takes as input the set of embeddings of the nodes in $\mathcal{N}(u)$ \cite{Hamilton2020}. When each node $u$ of the input graph has an associated $d_x$-dimensional input feature $\mathbf{x}_u \in \mathbb{R}^{d_x}$, $\mathbf{h}_u^{(0)}$ is set to $\mathbf{x}_u$. 
As a result, through several message-passing iterations $\mathbf{h}_u^{(k)}$ captures increasingly rich information encapsulating both the topological structure and the features surrounding each graph node $u$. However, after successive message-passing operation GNNs can suffer from vanishing gradients due to over-smoothing of the signal, leading to increasingly similar node representations \cite{Alon2021,Dwivedi2022a, Abboud2023}. In tasks where long-range interactions between far away nodes are important, this leads to loss of local neighborhood topological information.

\paragraph{Graph Attention Network.} Graph Attention Networks (GATs) \cite{Velickovic2018} are a type of GNN which incorporate masked self-attention layers \cite{Bahdanau2016,Vaswani2017} into message-passing and use attention weights to define a weighted sum of the neighbors, i.e.,

\begin{equation}
\mathbf{m}_{\mathcal{N}(u)}^{(k)}=\sum_{v \in \mathcal{N}(u)} \beta_{u, v} \mathbf{h}_v^{(k)} \,,
\end{equation}
where $\beta_{u, v}$ denotes the attention on neighbor $v \in \mathcal{N}(u)$ when aggregating information at node $u$. 

\paragraph{Graph pooling.} Graph pooling methods aim to downsample graphs while preserving essential structural information. There are two different type of approaches: spectral-based and top-$k$-based methods \cite{Ying2019}. Spectral approaches such as DiffPool \cite{Ying2019}, LaPool \cite{la_pool} or EigenPool \cite{eigenpool} transform the graph into a compressed representation through learned soft clustering assignments, producing new abstract node representations. In contrast, top-$k$ methods \cite{Zhang2019} such as gPool \cite{Gao2019a}, TopKPool \cite{Gao2019} or SAGPool \cite{sagpool} directly identify and preserve the most important nodes through various scoring mechanisms. The resulting scores enable direct node selection, maintaining a clear correspondence between the original and pooled graph, which maintains interpretability by producing a subgraph where node identity is conserved. gPool and TopKPool use a learnable vector to calculate projection scores and select the top-ranked nodes, but do not fully take into account graph topology \cite{sagpool,Cangea2018}. SAGPool \cite{sagpool} uses the GCN defined in \cite{Kipf2017} to calculate the self-attention scores $\mathbf{z} \in \mathbb{R}^{N \times 1}$ as follows:
\begin{equation}
\mathbf{z}=\sigma\left(\tilde{\mathbf{D}}^{-\frac{1}{2}} \tilde{\mathbf{A}} \tilde{\mathbf{D}}^{-\frac{1}{2}} \mathbf{X} \boldsymbol{\theta}_{a t t}\right),
\end{equation}
where $\tilde{\mathbf{A}} \in \mathbb{R}^{N \times N}$ represents the adjacency matrix with self-connections, $\tilde{\mathbf{D}}$ is its degree matrix, $\mathbf{X} \in \mathbb{R}^{N \times F}$ contains node features, and $\boldsymbol{\theta}_{a t t} \in \mathbb{R}^{F \times 1}$ are the learnable parameters.

By utilizing graph convolutions to obtain self-attention scores, the result of the pooling is based on both graph and topological features, while remaining efficient to calculate in terms of memory and runtime \cite{sagpool}. The node selection method follows \cite{Gao2019,Cangea2018,Knyazev2019} by retaining a portion of nodes of the input graph, even when graphs of varying sizes and structures are input. The pooling ratio $k \in(0,1]$ hyperparameter determines the number of nodes to keep at each pooling layer.

\paragraph{Graph readouts.} Graph readout operations are specifically focused on obtaining a fixed-size graph-level representation by aggregating all node features. This is generally done through simple statistical operators such as global mean and global max pooling operations \cite{Xu2019}. However, these basic aggregation procedures cause information loss through oversmoothing of the node signals, failing to capture complex topological relationships encoded into graphs. Recent methods have examined how to obtain more expressive graph readouts through the use of clustering \cite{clust_readout}, attention \cite{Chen2021a} or variance \cite{var_pool} based techniques. Notably in the histopathology area HEAT \cite{heat} proposed to aggregate based on the assignment of tissue type pseudo-labels. However, approaches based on pseudo-cluster can be inconsistent across graphs \cite{heat} and fail to align with meaningful and interpretable biology.

\paragraph{Positional encoding.}
Random walk positional encoding is a technique used to incorporate structural information from a graph into the node embeddings \cite{Dwivedi2022a}. Specifically, for each node $u$ in the graph, a random walk of fixed length is performed, starting from that node $u$ and considering only the landing probability of transitioning back to the node $u$ itself at each step, i.e., $
\mathbf{p}_{\text{RWPE}}^u = \begin{bmatrix}
RW_{uu}, RW^2_{uu}, \ldots RW^l_{uu}
\end{bmatrix}^\top \in \mathbb{R}^l \,,
$ where $\mathbf{p}_{\text{RWPE}}^u$ represents the random walk positional encoding for node $u$, $RW^l_{uu}$ is the $l$-step landing probability of returning to node $u$ after a random walk of length $l$ starting from $u$, and the positional encoding concatenates these $l$-step landing probabilities into a vector in $\mathbb{R}^l$. The node random walk positional encoding is then concatenated with its feature vector to obtain a new enriched input feature, i.e.,
$
\mathbf{h}_u = \mathbf{W}_c \left[ \mathbf{x}_u \, \Vert \, \mathbf{p}_{\text{RWPE}}^u \right] \,
$
where $\mathbf{h}_u \in \mathbb{R}^d$ is the final $d$-dimensional embedding for node $u$, $\mathbf{x}_u \in \mathbb{R}^{d_x}$ is the initial $d_x$-dimensional feature vector for node $u$, $\mathbf{p}_{\text{RWPE}}^u \in \mathbb{R}^l$ is the $l$-dimensional random walk positional encoding for node $u$, $\Vert$ denotes the vector concatenation operation, and $\mathbf{W}_c \in \mathbb{R}^{d \times (d_x + l)}$ is a learnable weight matrix that projects the concatenated node feature and positional encoding to an $d$-dimensional embedding space. This allows the node embeddings to capture not only the local neighborhood structure around each node, but also higher-order proximity information between nodes that are multiple hops away, potentially improving their ability to capture complex global patterns and dependencies within the graph structure. 
\section{Methods}
\label{sec:methods}

\begin{figure*}[!ht]
    \centering
    \includegraphics[width=1\linewidth]{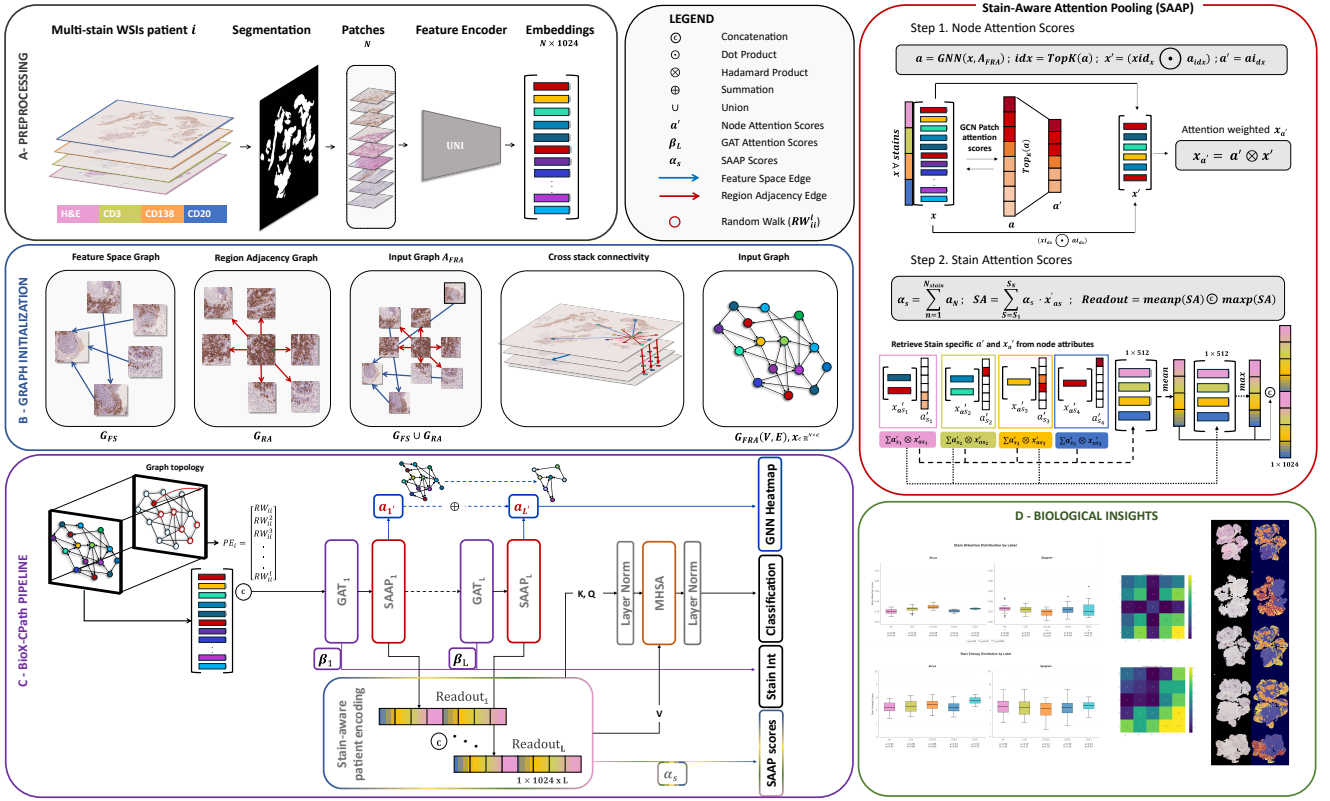}
    \caption{\textbf{Architecture}: Our approach begins by preprocessing the WSIs into patch features using UNI \cite{Chen2024a} (Section A). The resultant features are combined into two graphs, $G_{FS}$ and $G_{RA}$, representing the feature space similarity and region adjacency respectively. Given that the node sets of the two graphs are shared, we join the edge sets together, yielding graph $G_{FRA}$ (Section B). $G_{FRA}$ is then passed through hierarchical GNN blocks (Section C) consisting of a Graph Attention Network (GAT) \cite{Velickovic2018} and our proposed Stain-Aware Attention Pooling (SAAP) (detailed in the top right), which updates the node features while selecting the most relevant ones using an importance score. We obtain stain-aware patient encoding, which we pass through a final MHSA layer, before classification. Derived from both SAAP and GAT layers we propose metrics which provide biological insights into the model's predictions (Section D). }
    \label{fig:pipeline}
\end{figure*}

Here we introduce our proposed pipeline, which we illustrate graphically in Fig. \ref{fig:pipeline}. 

\subsection{Preprocessing}

\paragraph{Feature extraction.} We start by preprocessing each stack of patient multistain WSIs by thresholding tissue areas from background and extracting patches. For each extracted patch, the $(x,y)$-coordinates are saved. Each patch is then processed by a feature extractor to obtain an embedded feature vector. Here we use the UNI feature encoder \cite{Chen2024a} as it has shown reasonable performance on IHC benchmarking tasks \cite{AGS_2024}. This produces a feature matrix $\mathbf{X}_p \in \mathbb{R}^{N \times d}$ which represents the stack of WSIs for a given patient $p$, with $d$ the embedding dimension of the feature encoder. See Supplementary Material (SM).~\ref{technical_clarification} for further details.

\paragraph{Graph initialization.} Given our feature matrix $\mathbf{X}_p$, we first construct a $k$-Nearest Neighbor ($k$-NN) graph in feature space. This feature space graph $G_{FS}$ contains relationships between semantically similar patches, regardless of their spatial relationship, and has an adjacency matrix denoted $\mathbf{A}_{FS}[i,j]$ where $\mathbf{A}_{FS}[i,j] = 1$ if patch $j$ is among $k$ nearest neighbors of $i$ in feature space. We also construct a region adjacency graph $G_{RA}$ using the extracted $(x, y)$ coordinates, with adjacency matrix $\mathbf{A}_{RA}[i,j]$ where $\mathbf{A}_{RA}[i,j] = 1$ if patch $j$ is among $k$ region adjacent nearest neighbors of $i$ both on the $(x, y)$ plane (same WSIs) and $z$-axis of the WSIs stack. We illustrate these two types of connectivity in Fig. \ref{fig:pipeline}B. We then combine $\max(\mathbf{A}_{FS}, \mathbf{A}_{RA})$ to obtain our full $\mathbf{A}_{FRA} \in \{0,1\}^{N \times N}$, which we use to initialize our input graph $G_{FRA} = (V, E)$. For each node, we store as a categorical node attribute their stain type $S$, while for each edge we store the edge type. The combination of feature and spatial proximity was chosen to connect stains across the stack and permit information flow during message passing operations.  

\paragraph{Positional encoding.} For each node in $G_{FRA}$, a fixed length random walk is performed \cite{Dwivedi2022}, starting from a given node and considering only the landing probability of transitioning back to the initial node at each step. The random walk positional encoding vector is appended to the initial feature vector of its associated node and re-appended through each layer of our backbone. We employ this approach to alleviate issues with long-range cross WSIs stack connectivity by providing global topological information to the graph. 

% \paragraph{Stain Embedding} We also explore adding a fixed length learnable stain embedding appended to the initial feature vector. Each stain type is one-hot encoded to give a set of stain types $S={s_1, ..., s_n}$. These are then embedded using a learnable weight matrix $W_s$ to give a final Stain Embedding $(SE) = W_e \dot s_x$. 
 
\subsection{Patient Level Encoding}

\paragraph{Hierarchical graph blocks.}
To obtain patient-level encoding, we use as our backbone a hierarchal graph approach as presented in \cite{Cangea2018,Ying2019}, with the aim of attenuating oversmoothing issues. Our patient level encoder backbone consists of alternating GAT layers \cite{Velickovic2018} and our proposed Stain-Aware Attention Pooling (SAAP) module, which refines the node features whilst selecting the most relevant ones - using an importance score - to be forwarded to the next layer \cite{sagpool}. Finally, we apply multi-head self-attention (MHSA) to the concatenated stain-aware patients encoding returned by the SAAP module at each layer, with the resultant features passed to a fully connected classification head. Our backbone architecture choice is motivated by the desire to obtain the most expressive representation of patient encoding \cite{,h2_mil,graph_mil,AGS23_BMVC,heat}.

\paragraph{Stain-Aware Attention Pooling module.}

The SAAP algorithm, illustrated in Fig. \ref{fig:pipeline}, begins with calculating node attention scores $\mathbf{a} \in \mathbb{R}^N$. Here we use the SAGPool algorithm as described in Sec. \ref{sec:preliminaries}. Briefly, the attention scores are computed as
$\mathbf{a} = \text{GNN}(\mathbf{X}, \mathbf{A_{FRA}})$. These node attention scores represent the importance of each node in the graph based on both their features and graph topology. Both the node attention scores $\mathbf{a}$ and the feature matrix $\mathbf{X}$ are sorted and a subset $\mathbf{X}'$ is selected based on the top $k$ nodes wrt the attention scores, forming the subgraph $G'_{FRA}$. This preserves the most relevant nodes while reducing the computational complexity. The attention scores of the $k$ nodes are then used to scale the node features $\mathbf{X}'$ through element-wise multiplication.
%, i.e., $\mathbf{X'_{a'}} = \mathbf{X'} \otimes \mathbf{a}'$, where $\mathbf{a}'$ denote the attention scores of the top $k$ nodes.
This injects relevance ranking in the feature matrix $\mathbf{X}'$, such that more relevant nodes now have higher weight. For each stain $s$, a stain-level weight $\alpha_s$ is then calculated as the sum of the normalized attention scores $\mathbf{a}' = [a'_1, \cdots, a'_{Ns}]$ nodes belonging to that stain,

\begin{equation}
\alpha_s = \sum_{n=1}^{N_{s}} a'_n,
\end{equation}

The algorithm then pools weighted features by stain. The matrix of Stain Attention (SA) scores is calculated as 
\begin{equation}
    \text{SA scores} = \sum_{s \in S} \alpha_s \cdot \mathbf{X}'_s
\end{equation}
where $S$ is the set of stains, $\alpha_s$ is the attention weight for stain $s$ and $\mathbf{X}'_s$ represents the features specific to stain $s$. Finally, we obtain stain-aware readouts $\text{Readout} = [\text{meanp}(\text{SA}) \Vert \text{maxp}(\text{SA})]$
where $\text{meanp}$ and $\text{maxp}$ represent mean pooling and max pooling operations respectively, and $\Vert$ represents the vector concatenation operation. SAAP explicitly handles multiple stain modalities by computing stain-specific weights ($\alpha_s$), allowing the model to learn the relative importance of different stains for downstream tasks. With this we aim to maximize expressiveness, while aligning it with relevant biological information. 

\paragraph{Biological insights.} Based on our SAAP module and the proposed backbone architecture, we introduce a number of derived metrics which allow us to verify if the model aligns with known biology and can help provide clinical insights. These metrics are:

\begin{itemize}
    \item \textbf{SAAP scores}, defined above. This score informs us on which stains were most diagnostically relevant for the downstream task.
    \item \textbf{Stain entropy scores}, $H_s = -\sum_{n=1}^{N_s} (a'_n \cdot \log(a'_n))$
where $H_s$ is the entropy for stain $s$ and $a'_n$ are the normalized attention scores of the $N_s$ nodes belonging to stain $s$. This measures how uniformly distributed the attention scores are within each stain type, with lower entropy values indicating more concentrated, focused attention patterns aligning with organized, localized cellular structures, while higher entropy represents uniformly distributed attention corresponding to diffuse, disorganized cellular structures present throughout the tissue.
    \item \textbf{Stain-stain interaction scores}, $\mathbf{I}$ are defined as $\mathbf{I}_{i,j} = \mathbf{I}_{j,i} = \frac{1}{|P_{i,j}|} \sum_{p \in P_{i,j}} \beta_p$
where $i,j$ are indices in the set of unique stains $S$, $P_{i,j}$ is the set of all pairs between stains $s_i$ and $s_j$ and $\beta_p$ represents the GAT attention weights for pair $p$, extracted from the model's attention mechanism. $|P_{i,j}|$ is the number of pairs between stains $s_i$ and $s_j$. This score quantifies the importance of edge connections between nodes of different stain types.
\end{itemize}

\paragraph{GNN Heatmap.} Extending on the use of attention scores, we design a simple GNN heatmap visualization method. The attention scores calculated for each node at the first SAAP layer are extracted and successively updated after each the pooling procedure. The final attention scores are min-max normalized and mapped back to their spatial location to obtain an attention heatmap of node importance. The resulting heatmap overlay provides a visual interpretation of the GNN model attention, highlighting the regions of the image that are considered most important for the downstream task.
\section{Experiments}
\label{sec:experiments}

\subsection{Datasets}

We test our pipeline on two autoimmune multi-stain datasets, one for Rheumatoid Arthritis and the other for Sjogren's Disease. Each dataset is composed of H\&E slides, with approximately 3 IHC slides of different immune biomarkers per patient. In SM.~\ref{app:ihc}, we give further information on the stains present in each dataset. 

\subsubsection{Rheumatoid Arthritis}

This dataset consists of 607 WSIs from 153 RA patients, categorized into low (N=66) and high (N=87) inflammatory subtypes \cite{Humby2021}. Samples were stained with H\&E and the IHC markers CD20+ B cells, CD68+ macrophages, and CD138+ macrophages (see Fig. SM.~\ref{ra_stain_types}). The dataset features a variable number of stains, averaging 3.9 per patient. We perform binary classification on low (N=66) and high (N=87) inflammatory subtypes. We extract non-overlapping patches at a 10x magnification, keeping those with over 40\% tissue coverage, totaling approximately 275k patches.

\subsubsection{Sjogren} This dataset consists of 347 WSIs labial salivary gland biopsies sampled from 93 patients, with 46 cases of non-specific Sicca and 47 cases of Sjogren. Samples were stained with H\&E and the IHC stains CD20+ B cells, CD3+ T cells, and CD138+ plasma cells, as well as CD21+ follicular dendritic cell network, only when B-cell aggregates were shown by the CD20+ staining for suspected cases of Sjogren (see Fig. SM.~\ref{sd_stain_types}). Each patient has a variable set of multi-stain WSIs, averaging 3.7 stains per patient. We perform detection of inflammatory patterns. We extract non-overlapping patches at a 20x magnification, keeping those with over 30\% tissue coverage, totaling approximately 237k patches. 

\subsection{Implementation Details} 

\quad \textbf{Experimental setup and evaluation metrics.} We separate a random label stratified 20\% hold out test set and perform 5-fold random label stratified cross-validation on the remaining data (train:val:test / 60:20:20). Models were trained for a maximum 200 epochs, with patience set to 15 such that early stopping was called if no change was observed in either the loss, accuracy, or AUC score for 15 epochs. Weights were kept for the model obtaining the best accuracy score on each validation set while ensuring there was no under-fitting or over-fitting of the models. Each of the 5 trained models was applied to the hold-out test. We report the mean and standard error (SE) of the results obtained on the hold-out test set for accuracy, macro F1-score, precision, recall, AUC, and average precision.

\textbf{Training schedule.} All models were trained using cross-entropy loss, with the AdamW optimizer set to $\beta_1=0.9$, $\beta_2=0.98$, and $\epsilon=10^{-9}$, with a learning rate $1 e^{-3}$ and weight decay $L_2=0.01$. No learning scheduler was used. We show the hyperparameters used in Table SM.3. Training was conducted on an NVidia A100 GPU (40Gb). See SM. \ref{hyperparameters}, SM.~\ref{memory_usage} for hyperparameters used and peak VRAM and memory use.

\textbf{Benchmarking and ablation studies.} We compare our method against seven SOTA methods, ABMIL \cite{Ilse18}, CLAM-SB \cite{Lu2021}, DeepGraphConv \cite{Li2018} PatchGCN \cite{Chen2021a}, TransMIL \cite{shao2021transmil}, GTP \cite{Zheng2022} and MUSTANG \cite{AGS23_BMVC}. We perform ablation on the different components of our pipeline: the SAAP module, the RW positional encoding, and the Multi Head Self-Attention layer.

\section{Results}
\label{sec:results}

In Table \ref{tab:main_results} we present the results obtained by BioX-CPath on both datasets. On the RA dataset, our model achieved 0.90 ($\pm$0.019) accuracy, representing a 4 percent point improvement over the next best performing model, MUSTANG (0.86 $\pm$0.021). BioX-CPath did not outperform MUSTANG in AUC (0.96 $\pm$0.007) and average precision (0.98 $\pm$0.004), however did well compared to other methods. On the Sjogren dataset BioX-CPath achieved 0.84 ($\pm$0.018) accuracy, showing a significant improvement over both CLAM-SB and MUSTANG (0.80 $\pm$0.018). The model also demonstrated stronger AUC (0.88 $\pm$0.023) and average precision (0.86 $\pm$0.032) compared to all baseline methods.

Ablation results shown in Tables \ref{sjogren_ablation} and \ref{ra_ablation} highlight the contribution of each component in our model. On the Sjogren dataset, the baseline model achieved 0.756 ($\pm$0.059) accuracy, while adding the RW positional encoding improved the performance to 0.80 ($\pm$0.038), indicating the importance of adding long-range topological information to the graph. The addition of SAAP provided another substantial boost, bringing the accuracy to 0.84 ($\pm$0.018). Similarly for the RA dataset, while the positional encoding improved the accuracy from 0.79 ($\pm$0.018) to 0.86 ($\pm$0.018), the full model with SAAP achieved the best performance at 0.90 ($\pm$0.019). We note the addition of the MHSA brought a slight decrease in performance. However, given the gains in model interpretability we do not view this as a significant disadvantage (we discuss this further in SM.8). 

\begin{table*}[]
\centering
\caption{Performance comparison of BioX-CPath against SOTA methods on the RA and Sjogren datasets. We report accuracy, AUC, and average precision (AP) with standard error shown in parentheses. The best results for each metric are shown in bold, with the second best underlined.}
\label{tab:main_results}
\resizebox{0.9\textwidth}{!}{%
\begin{tabular}{rcccccc}
\toprule
\multicolumn{1}{l}{} & \multicolumn{3}{c}{\textbf{RA}} & \multicolumn{3}{c}{\textbf{Sjogren}} \\ \midrule
\multicolumn{1}{l}{} & \textbf{Accuracy ($\uparrow$)} & \textbf{\textbf{AUC} ($\uparrow$)} & \textbf{AP} ($\uparrow$) & \textbf{Accuracy ($\uparrow$)} & \textbf{\textbf{AUC} ($\uparrow$)} & \textbf{AP} ($\uparrow$) \\
\textbf{ABMIL} \cite{Ilse18} & 0.79 {\scriptsize(0.028)} & 0.89 {\scriptsize(0.027)} & 0.92 {\scriptsize(0.019)}  & 0.73 {\scriptsize(0.018)} & 0.80 {\scriptsize(0.035)} & 0.79 {\scriptsize(0.044)} \\
\textbf{CLAM-SB} \cite{Lu2021} & 0.81 {\scriptsize(0.026)} & 0.92 {\scriptsize(0.011)} & 0.95 {\scriptsize(0.008)} & \underline{0.80 {\scriptsize(0.018)}} & \underline{0.85 {\scriptsize(0.017)}} & 0\underline{.85 {\scriptsize(0.026)}} \\
\textbf{TransMIL} \cite{shao2021transmil} & 0.80 {\scriptsize(0.025)} & 0.87 {\scriptsize(0.024)} & 0.91 {\scriptsize(0.021)} & 0.75 {\scriptsize(0.018)} & 0.73 {\scriptsize(0.011)} & 0.74 {\scriptsize(0.017)} \\
\textbf{DeepGraphConv} \cite{Li2018} & 0.81 {\scriptsize(0.025)} & 0.88 {\scriptsize(0.009)} & 0.92 {\scriptsize(0.007)} & 0.77 {\scriptsize(0.038)} & 0.83 {\scriptsize(0.031)} & 0.83 {\scriptsize(0.039)} \\
\textbf{Patch-GCN} \cite{Chen2022} & 0.83 {\scriptsize(0.015)} & 0.91 {\scriptsize(0.019)} & 0.94 {\scriptsize(0.014)} & 0.77 {\scriptsize(0.019)} & 0.85 {\scriptsize(0.015)} & 0.83 {\scriptsize(0.030)} \\
\textbf{GTP} \cite{Zheng2022} & 0.79 {\scriptsize(0.020)} & 0.87 {\scriptsize(0.012)} & 0.92 {\scriptsize(0.007)} & 0.62 {\scriptsize(0.048)} & 0.73 {\scriptsize(0.031)} & 0.72 {\scriptsize(0.024)} \\
\textbf{MUSTANG} \cite{AGS23_BMVC} &\underline{ 0.86 {\scriptsize(0.021)}} & \underline{0.96 {\scriptsize(0.010)}} & \underline{0.97 {\scriptsize(0.006)}} &\underline{ 0.80 {\scriptsize(0.018)}} & \underline{0.85 {\scriptsize(0.019)}} & 0.84 {\scriptsize(0.026)} \\
\textbf{BioX-CPath [ours]} & \textbf{0.90 {\scriptsize(0.019)}} & \textbf{0.96 {\scriptsize(0.007)}} & \textbf{0.98 {\scriptsize(0.004)}} &\textbf{ 0.84 {\scriptsize(0.018)}} & \textbf{0.88 {\scriptsize(0.023)}} & \textbf{0.86 {\scriptsize(0.032)}} \\ \bottomrule
\end{tabular}
}
\end{table*}

\begin{table}[]
\centering
\caption{Ablation on model components shown on the Sjogren dataset.}
\label{sjogren_ablation}
\resizebox{0.8\columnwidth}{!}{%
\begin{tabular}{@{}lccc@{}}
\toprule
  & \textbf{Accuracy ($\uparrow$)} & \textbf{AUC ($\uparrow$)} & \textbf{AP ($\uparrow$)} \\ \midrule
\textit{Baseline} & 0.756 (0.059) & 0.849 (0.024) & 0.84 (0.036) \\
+ \textit{MHSA} & 0.736 (0.049) & 0.849 (0.021) & 0.86 (0.036) \\
+ \textit{RW}  & 0.80 (0.038) & 0.84 (0.035) & 0.81 (0.034)  \\
\toprule
+ \textit{SAAP}  &\textbf{ 0.84 (0.018)} & \textbf{0.88 (0.023)} & \textbf{0.86 (0.032)} \\
\bottomrule
\end{tabular}}
\end{table}

% \usepackage{graphicx}
\begin{table}[]
\centering
\caption{Ablation on model components shown on the RA dataset.}
\label{ra_ablation}
\resizebox{0.8\columnwidth}{!}{%
\begin{tabular}{lccc}
\toprule
 & \textbf{Accuracy ($\uparrow$)} & \textbf{AUC ($\uparrow$)} & \textbf{AP ($\uparrow$)} \\ \midrule
\textit{Baseline} & 0.79 (0.018) & 0.87 (0.011) & 0.92 (0.010) \\
+ \textit{MHSA} & 0.78 (0.025) & 0.88(0.024) & 0.92 (0.018) \\
+ \textit{RW} & 0.86 (0.018) & 0.95 (0.010) & 0.98 (0.007) \\ 
\toprule
+ \textit{SAAP} & \textbf{0.90 (0.019)} & \textbf{0.96 (0.007)} & \textbf{0.98 (0.004)} \\ 
\bottomrule
\end{tabular}%
}
\end{table}

\subsection{Biological Interpretability}

\subsubsection{RA}

\begin{figure}
\centering
    \includegraphics[width=1\linewidth]{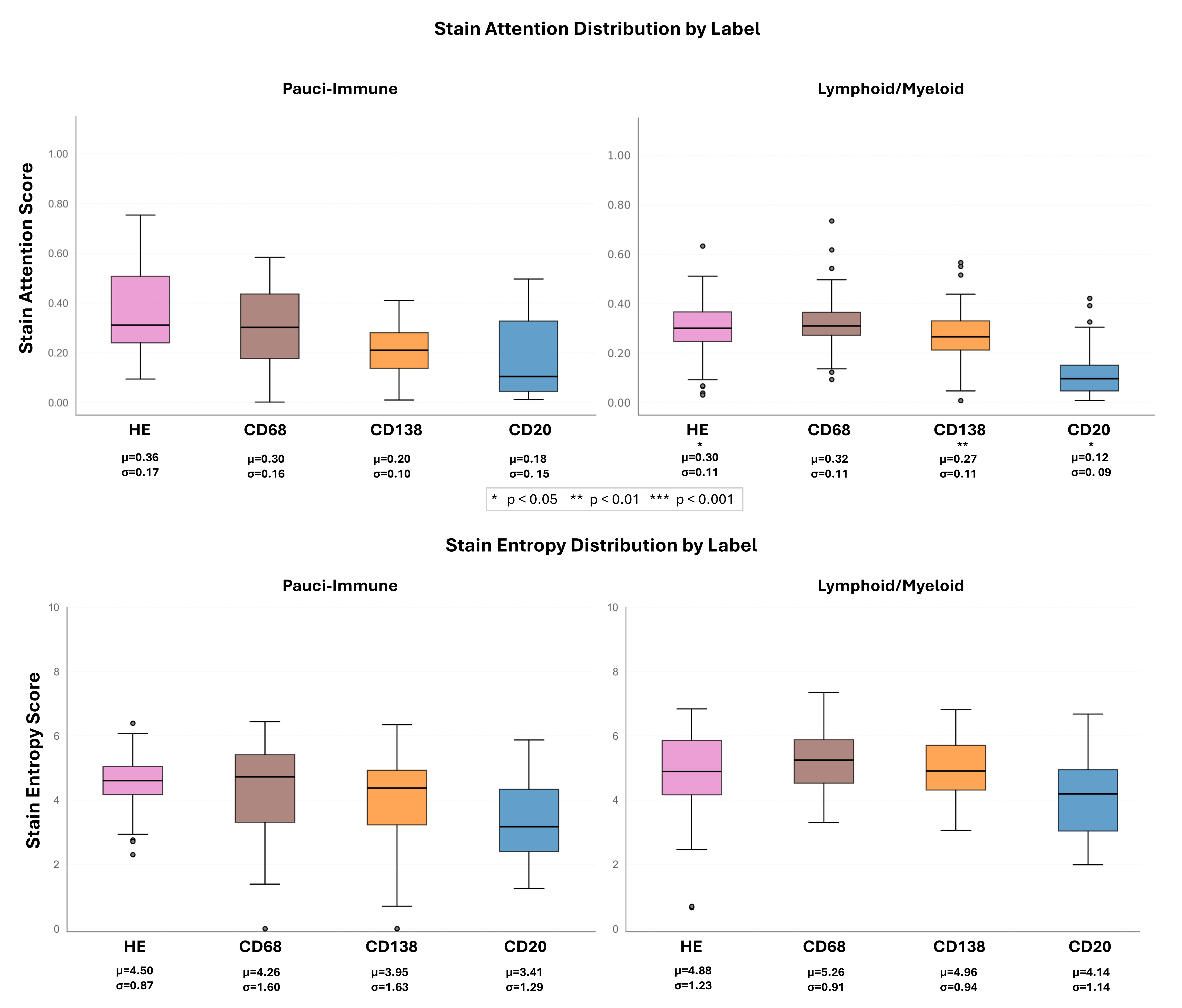}
\caption{\textbf{RA Dataset Explainability}: The top row shows box plots of the SAAP scores distribution for different stain types (H\&E, CD68, CD138, and CD20) for each classification label in the RA dataset (Pauci-Immune and Lymphoid/Myeloid). The bottom row shows the entropy score distributions for each of the stain types according to the classification label.}
\label{fig:stain_ent_ra}
\end{figure}

The Pauci-Immune pathotype exhibited lower attention scores for CD138 ($\mu=0.20$, $\sigma=0.10$, $p<0.01$) and CD20 ($\mu=0.18$, $\sigma=0.15$, $p<0.05$) markers, reflecting the characteristic scarcity of lymphocytic and plasma cell infiltrates in this disease subset. The lower entropy values observed in these samples (CD20: $\mu=3.41$, $\sigma=1.29$; CD138: $\mu=3.95$, $\sigma=1.63$) quantitatively capture the more ordered tissue architecture and sparse inflammatory foci associated with this RA pathotype. Conversely, Lymphoid/Myeloid samples showed more balanced attention distribution across CD68 ($\mu=0.32$, $\sigma=0.11$) and CD138 ($\mu=0.27$, $\sigma=0.11$) with consistently higher entropy values (CD68: $\mu=5.26$, $\sigma=0.91$; CD138: $\mu=4.96$, $\sigma=0.94$), reflecting the established role of plasma cells and macrophages in driving severe disease through autoantibody production and pro-inflammatory cytokine secretion \cite{Dennis2014,zhang2019_ra}. These computational findings provide quantitative support for the histological classification of RA subtypes, where Lymphoid/Myeloid pathotypes demonstrate abundant but disorganized immune cell infiltrates, while Pauci-Immune samples show more limited inflammatory patterns and more ordered tissue architecture \cite{humby2019synovial,Lewis2019}. The relatively high attention scores for H\&E staining in Pauci-Immune ($\mu=0.36$, $\sigma=0.17$, $p<0.05$) align with the understanding that when specific immune cell infiltrates are less prominent, general tissue architecture becomes more informative for pathotype classification, reflecting the heterogeneous nature of RA synovitis and its immunological basis.

\subsubsection{Sjogren}

\begin{figure}
\centering
    \includegraphics[width=1\linewidth]{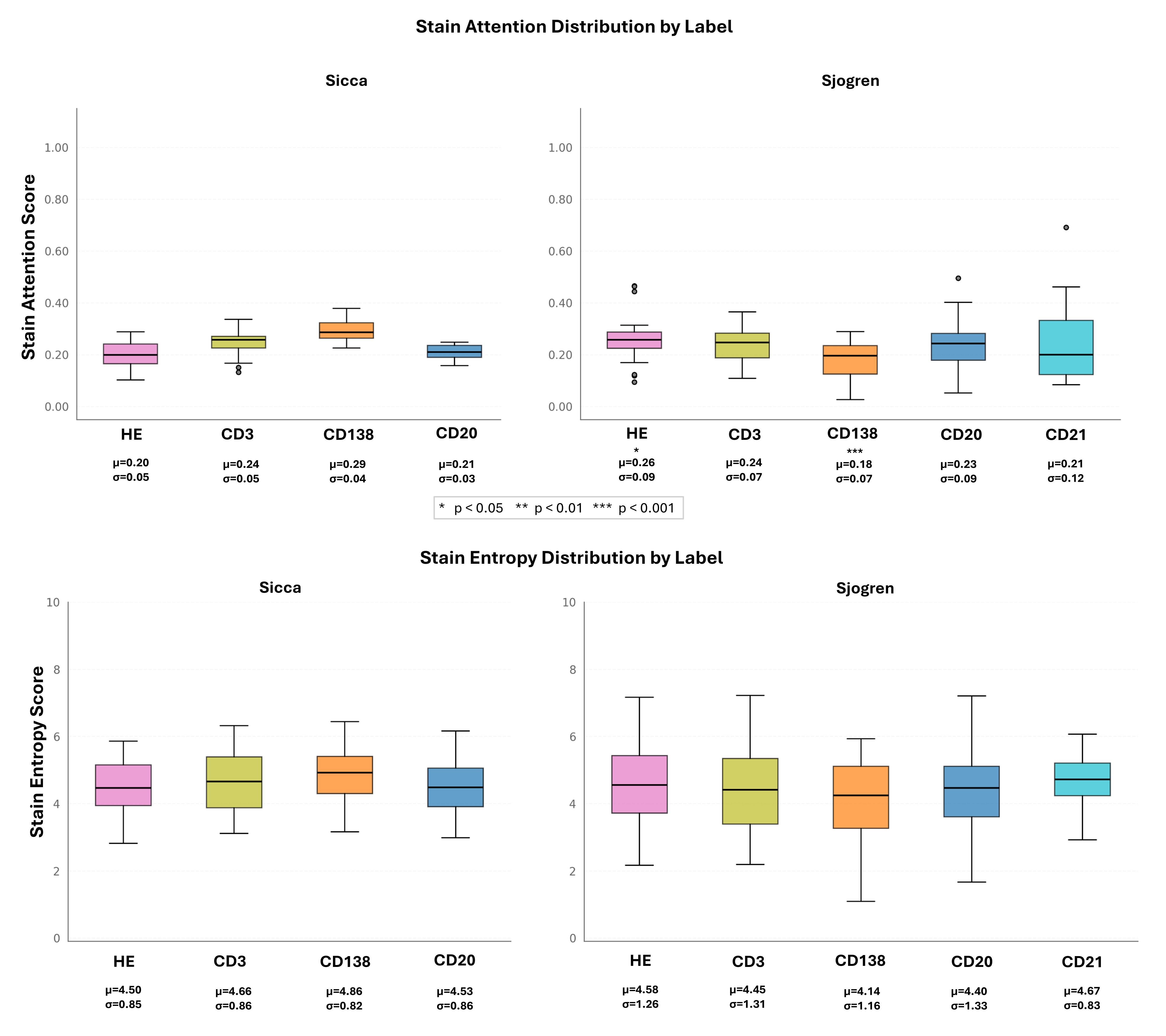}
\caption{\textbf{Sjogren Dataset Explainability}: The top row shows box plots of the SAAP scores for different stain types (HE, CD3, CD138, CD20, and CD21) for each classification label in the Sjogren dataset (Sicca and Sjogren). The bottom row shows the entropy score distributions for each of the stain types according to the classification label.}
\label{fig:stain_ent_sd}
\end{figure}

Sjogren's samples show a balanced attention across immune markers, with CD3 ($\mu=0.24$, $\sigma=0.07$), CD20 ($\mu=0.23$, $\sigma=0.09$), and CD21 ($\mu=0.21$, $\sigma=0.12$) receiving balanced attention, along with HE staining ($\mu=0.26$, $\sigma=0.09$). This reflects characteristic organized lymphocytic infiltrates with a mix of B-cells, plasma cells, and T-cells \cite{bombardieri2017}, as well as the importance of changes in tissue architecture. Most notably CD138 shows significantly lower attention in the Sjogren's group ($\mu=0.18$, $\sigma=0.07$) compared to Sicca ($\mu=0.29$, $\sigma=0.04$, $p<0.001$), with lower entropy scores ($\mu=4.14$, $\sigma=1.16$) suggesting that specific plasma cell organization patterns, rather than overall abundance, are distinctive for Sjogren's pathology \cite{pontarini2024}. These patterns align with current understanding where Sicca represents non-inflammatory dryness with more homogeneous tissue (higher entropy), while Sjogren's demonstrates organized autoimmune infiltrates with more concentrated immune cell groupings (lower entropy). The model appears to have learned biologically relevant features that align with known pathological mechanisms. \\

In Supplementary materials, we conduct further analysis of model interpretability, looking at stain interaction scores (SM.~\ref{stain_int}), GNN heatmaps (SM.~\ref{gnn_heatmap}) and Layer Importance (SM.~\ref{layer_importance}).

\section{Conclusion}
\label{sec:conclusion}

BioX-CPath is an explainable GNN-based architecture for multistain IHC analysis, that bridges computational pathology and biological interpretability. By integrating multistain histopathological data into a unified framework, our approach not only achieves state-of-the-art accuracy but also provides mechanistic insights that align with established pathological mechanisms. This work establishes a foundation for developing and extending explainable computational pathology to other complex autoimmune and inflammatory diseases where multistain tissue analysis is essential for accurate diagnosis and subtyping.

{\section*{Acknowledgments and Disclosure of Funding}

We wish to thank Dr. Dovile Zilenaite for her insightful comments and knowledge, in particular discussing stain-stain interaction and entropy scores. A.G.S. receives funding from the Wellcome Trust [218584/Z/19/Z]. This paper utilized Queen Mary’s Andrena HPC facility \cite{King2017}. This work also acknowledges the support of the National Institute for Health and Care Research Barts Biomedical Research Centre (NIHR203330), a delivery partnership of Barts Health NHS Trust, Queen Mary University of London, St George’s University Hospitals NHS Foundation Trust and St George’s University of London.}

% {\small
% \bibliographystyle{ieeenat_fullname}
% \bibliography{krag}
% }

\clearpage
\setcounter{page}{1}
\maketitlesupplementary
\appendix

The content of the supplementary material are as follows: in \ref{app:ihc} we describe IHC and the different immune markers used in this study and in \ref{dataset}, we give further details on the datasets. We go over hyperparameters, memory usage and further technical clarification in sections \ref{hyperparameters}, \ref{memory_usage} and \ref{technical_clarification}. Finally, we conduct further analysis of model interpretability, by looking at stain interaction scores in \ref{stain_int}, GNN heatmaps in \ref{gnn_heatmap} and Layer Importance \ref{layer_importance}.

\section{Immunohistochemistry Staining}
\label{app:ihc}

IHC serves as a critical molecular mapping tool in clinical diagnostics and research, enabling precise identification and localization of disease-specific markers. The technique's power lies in its ability to reveal the molecular and cellular landscape of pathological processes, providing crucial information for diagnosis, prognosis, and treatment decisions.

In autoimmune disease diagnosis and monitoring, IHC enables detailed immune cell profiling through the characterization of inflammatory infiltrates and quantification of specific immune cell populations. This information reveals patterns of autoantibody deposits, complement activation, and tissue-specific autoantigen expression. The technique proves particularly valuable in assessing disease activity through the evaluation of inflammatory marker expression and monitoring tissue damage and repair processes.

IHC's integration into clinical decision-making represents a cornerstone of modern pathology practice. It supports diagnostic algorithms by validating initial morphological findings and resolving differential diagnoses through confirmation of disease-specific molecular patterns. In treatment strategy development, IHC helps identify targetable pathways and predict treatment response, enabling more personalized therapeutic approaches. 

\subsection{CD Markers}

CD markers (Cluster of Differentiation) are cell surface proteins that serve as essential identifiers in immunological analysis. Each marker identifies specific immune cell types, enabling detailed characterization of tissue immune responses.

\begin{itemize}
    \item \textbf{CD20} is a B-lymphocyte-specific antigen expressed on the surface of pre-B and mature B cells. This marker is critically important in both diagnostic and therapeutic contexts, particularly in B-cell lymphomas and autoimmune disorders. CD20 serves as the target for rituximab and other monoclonal antibody therapies, making its detection crucial for treatment planning. In lymphoid tissue analysis, CD20 staining helps identify B-cell populations and assess their distribution within tissue architecture.
    \item \textbf{CD21} is predominantly expressed on mature B cells and follicular dendritic cells. It plays a crucial role in the formation and maintenance of germinal centers within lymphoid tissues. In diagnostic pathology, CD21 staining is particularly valuable for visualizing follicular dendritic cell networks and assessing lymphoid tissue organization. This marker is often used to evaluate lymphoid tissue architecture in conditions such as lymphomas and autoimmune disorders.
    \item \textbf{CD68} is a glycoprotein expressed primarily by macrophages and monocytes. In tissue analysis, CD68 serves as a reliable marker for identifying tissue-resident macrophages and assessing inflammatory responses. In autoimmune disease diagnostics, CD68 staining helps quantify macrophage infiltration and assess disease activity.
    \item \textbf{CD138} is a transmembrane heparan sulfate proteoglycan primarily expressed on plasma cells and some epithelial cells. In autoimmune disease diagnostics, CD138 helps evaluate plasma cell infiltration and potential antibody production sites within affected tissues.
    \item \textbf{CD3} is a fundamental marker of T lymphocytes, expressed throughout T-cell development and maintained on mature T cells. CD3 staining is crucial in diagnosing T-cell lymphomas, assessing T-cell-mediated immune responses. In the context of autoimmune diseases, CD3 staining helps characterize the T-cell component of inflammatory infiltrates.
\end{itemize}

These markers, when analyzed together, map the immune cell landscape within tissues, revealing patterns of immune response and inflammation that guide diagnosis and treatment decisions. 

\section{Dataset Characteristics}
\label{dataset}

\begin{figure}[!h]
\centering
\includegraphics[width=\columnwidth]{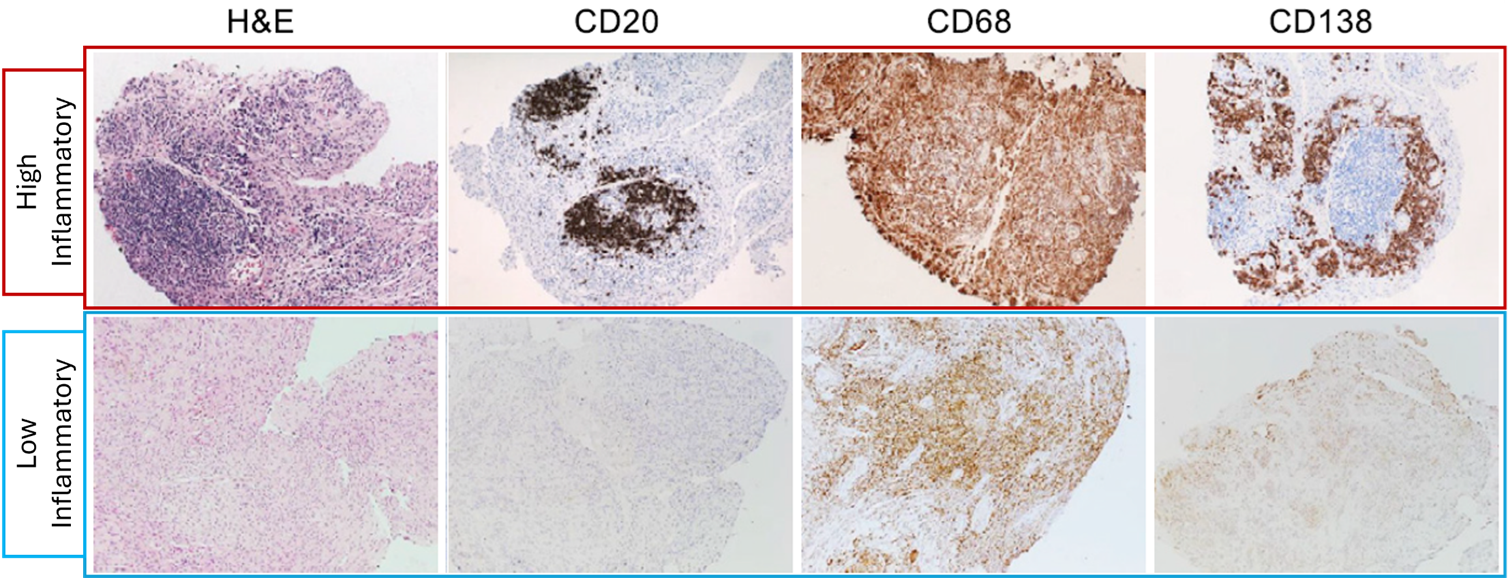}
\caption{\textbf{Example of low inflammatory vs high inflammatory pathotype presentation in H\&E and IHC stains for RA}: Rheumatoid Arthritis inflammatory pathotypes based on semi-quantitative analysis of synovial tissue biopsies stained with H\&E, CD20+ B cells, CD68+ macrophages and IHC+ CD138 plasma cells.}
\label{ra_stain_types}
\end{figure}

\begin{figure}[!h]
\centering
\includegraphics[width=\columnwidth]{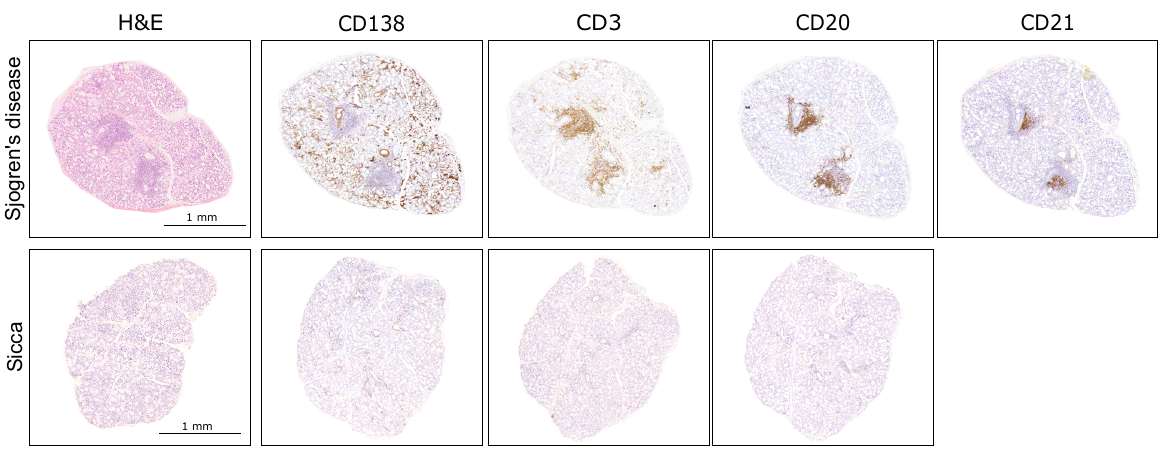}
\caption{\textbf{Example of Sicca vs Sjogren presentation in H\&E and IHC stains}: On bottom, a patient diagnosed with Sicca, on top a patient diagnosed with Sjogren's Disease. Here we show samples stained with IHC stains CD3+ T cells, CD20+ B cells, and CD138+ plasma cells, as well as CD21+ in the case of Sjogren's Disease.}
\label{sd_stain_types}
\end{figure}

To provide a benchmark on autoimmune multistain datasets, we use two clinical datasets. One dataset derives from a clinical trial, where patients with difficult to treat RA were recruited for treatment with rituximab. The other dataset derives from WSIs gathered for research purposes with the purpose of examining differences between patients presenting with dry eyes and mouth (Sicca) and patients subsequently diagnosed with Sjogren's Disease. In Figs. \ref{ra_stain_types} and \ref{sd_stain_types}, we present clear examples of RA pathotypes and Sicca versus Sjogren presentation. While these images highlight characteristic differences, they represent more extreme cases specifically selected for illustrative clarity. The actual dataset exhibits considerably more heterogeneity in presentation, with many cases showing more subtle differences. In Table \ref{metadata}, we give further information on the stains present in each dataset. Each dataset is composed of H\&E slides, with approximately 3 IHC slides of different immune biomarkers per patient. 

\begin{table}[!h]
 \centering
\caption{\textbf{Metadata and dataset characteristics} for Sjogren and RA cohorts, including number of patients, WSIs, stains present and average number of stains per patient. We highlight in pink {\color[HTML]{AA336A}H\&E} staining and blue {\color[HTML]{00009B}IHC}.}
\label{metadata}
\resizebox{0.9\columnwidth}{!}{%
\begin{tabular}{@{}rcccc@{}}
\toprule
\multicolumn{1}{l}{} & \multicolumn{2}{c}{\textbf{Sjogren}} & \multicolumn{2}{c}{\textbf{Rheumatoid Arthritis}} \\ \midrule
\textbf{No. Patients} & \multicolumn{2}{c}{93} & \multicolumn{2}{c}{153} \\
\textbf{No. Slides} & \multicolumn{2}{c}{347} & \multicolumn{2}{c}{607} \\
\textbf{No. Stains} & \multicolumn{2}{c}{5} & \multicolumn{2}{c}{4} \\
\textbf{Av. Stains per patient} & \multicolumn{2}{c}{3.7} & \multicolumn{2}{c}{3.97} \\
\textbf{Magnification} & \multicolumn{2}{c}{20x} & \multicolumn{2}{c}{10x} \\
\textbf{Total no. patches} & \multicolumn{2}{c}{237k} & \multicolumn{2}{c}{275k} \\
\textbf{Av. Patches per patient} & \multicolumn{2}{c}{2 530} & \multicolumn{2}{c}{1800} \\ \midrule
\textbf{Patches per stain} & \textbf{Mean} & \textbf{Total} & \textbf{Mean} & \textbf{Total} \\
{\color[HTML]{AA336A} \textbf{HE}} & 650 & 61055 & 434 & 66511 \\
{\color[HTML]{00009B} \textbf{CD3}} & 625 & 58712 & 0 & 0 \\
{\color[HTML]{00009B} \textbf{CD138}} & 377 & 35416 & 481 & 73624 \\
{\color[HTML]{00009B} \textbf{CD20}} & 626 & 58805 & 351 & 53768 \\
{\color[HTML]{00009B} \textbf{CD21}} & 254 & 23843 & 0 & 0 \\
{\color[HTML]{00009B} \textbf{CD68}} & 0 & 0 & 535 & 81915 \\ \midrule
\textbf{ML problem type} & \textbf{Detection} & \textbf{} & \textbf{Subtyping} &  \\
\textbf{Labels} & Negative & 46 & \begin{tabular}[c]{@{}c@{}}Low \end{tabular} & 66 \\
\multicolumn{1}{c}{} & Positive & 47 & \begin{tabular}[c]{@{}c@{}}High \end{tabular} & 87 \\ \bottomrule
\end{tabular}%
}
\end{table}

\section{Hyperparameters}
\label{hyperparameters}

We trained using the AdamW optimizer set to $\beta_1=0.9$, $\beta_2=0.98$ and $\epsilon=10^{-9}$, with a learning rate $1 e^{-3}$ and weight decay $L_2=0.01$. No learning scheduler was used. We show our model's hyperparameters in Table \ref{tab:hyperparameters}. 

\begin{table}[!h]
\centering
 \caption{\textbf{Our model hyperparameters}. We provide the hyperparameters used for each dataset to train our model.}
 \label{tab:hyperparameters}
\resizebox{0.9\columnwidth}{!}{%
\begin{tabular}{llllllll}
\toprule
Dataset &Seed & LR & \# Layers & PE Dim & Pooling Ratio & Attention Heads & Dropout \\
\midrule
RA & 42 & 0.0001 & 4 & 20 & 0.7 & 2 & 0.2 \\
Sjogren & 42 & 0.0001 & 4 & 20 & 0.5 & 4 & 0.2 \\
\midrule
\end{tabular}
}
\end{table}

\section{Memory Usage}
\label{memory_usage}

Table \ref{tab:memory-usage} presents the RAM and VRAM utilization across all models compared against BioX-CPath. The varying RAM requirements stem from the distinct input representations each model processes: ABMIL/CLAM/TransMIL operate on embeddings, PatchGCN/GTP utilize region adjacency graphs, DeepGraphConv and MUSTANG work with feature space graphs, while BioX-CPath processes both feature and region adjacency graphs. VRAM consumption differences reflect the architectural complexity of each model. While simpler architectures like ABMIL \cite{Ilse18} demonstrate minimal VRAM usage, our model's incorporation of GAT self-attention operations and an additional MHSA mechanism for interpretability results in higher peak VRAM consumption. We consider this increased memory footprint an acceptable trade-off given the model's superior performance and enhanced explainability. Future research could focus on developing a more memory-efficient architecture that maintains these characteristics, enabling translation to clinical practice.

\begin{table}[h!]
 \centering
 \caption{\textbf{Training and inference memory usage}. The table shows both RAM and VRAM peak usage during training and inference for the benchmark models shown in the main results table. We present results for the Sjogren dataset. Lower is \textbf{better}.}
 \label{tab:memory-usage}
\resizebox{1\columnwidth}{!}{%
\begin{tabular}{lllll}
\toprule
\multirow{2}{*}{Model} & \multicolumn{2}{l}{Training}                  & \multicolumn{2}{l}{Inference}                \\ \cline{2-5} 
                       & RAM (GB $\downarrow$) & VRAM (GB$\downarrow$) & RAM (GB$\downarrow$) & VRAM (GB$\downarrow$) \\
\midrule
ABMIL \cite{Ilse18} & 38.11 & \textbf{0.09} & 32.93 & \textbf{0.09} \\
CLAM-SB \cite{Lu2021} & 44.38 & 0.14 & 45.03 & 0.10 \\
TransMIL \cite{shao2021transmil} & \textbf{35.87} & 1.47 & \textbf{29.39} & 0.79 \\
DeepGraphConv \cite{Li2018} & 55.65 & 1.31 & 45.10 & 0.68 \\
Patch-GCN \cite{Chen2022} & 41.03 & 7.42 & 41.99 & 4.37 \\
GTP \cite{Zheng2022} & 47.11 & 2.40 & 48.15 & 1.97 \\
MUSTANG \cite{AGS23_BMVC} & 36.00 & 6.18 & 36.25 & 3.52 \\
BioX-CPath (ours) & 41.30 & 11.25 & 36.61 & 9.19 \\
\midrule
\end{tabular}
}
\end{table}

\section{Technical clarification}
\label{technical_clarification}

The feature matrix is obtained through a hierarchical data loading architecture, which enables us to deal with the variable number of stains per patient: (1) A slide-level DataLoader processes each stain-specific WSI, extracting patches and associated metadata (stain type, spatial coordinates, patient ID); (2) A patient-level loader stacks the stain-specific embeddings through vertical concatenation; (3) graphs are constructed using patch embeddings as node features with dual-criteria edge connectivity (feature and spatial proximity). The preprocessed patient graphs are then stored, loaded \& batched with PyTorch Geometric DataLoader. We keep track of node and edge attributes, stored as categorical labels, through each layer of our model by mapping and storing their IDs after each pooling operation. When nodes are removed, edges are systematically pruned where either the source or target node was dropped, updating the edge list accordingly. While this can lead to disconnected components, the high initial connectivity of the patient graphs means these components emerge only in deeper layers of the encoder, where they exhibit ``specialized" attention patterns focusing on specific stain or tissue regions. We exemplify this with a layer-wise graph WSIs overlay shown in Figure~\ref{overlay}. The max (OR) operator was chosen over min (AND), based on graph connectivity patterns: using AND overly restricts edges ($\sim$10\%), limiting message passing and cross-stain interactions. In contrast, OR preserves local and global connectivity, allowing the SAAP module to dynamically prioritize relevant edges. These design choices are all aimed at optimizing computational resources and information flow, under minimal supervision requirements (patient-level labels and stain-type slide annotations), while ensuring interpretable biologically-aligned results.

\section{Stain-Stain Interactions}
\label{stain_int}

\begin{figure*}[!t]
    \centering
    \includegraphics[width=1\linewidth]{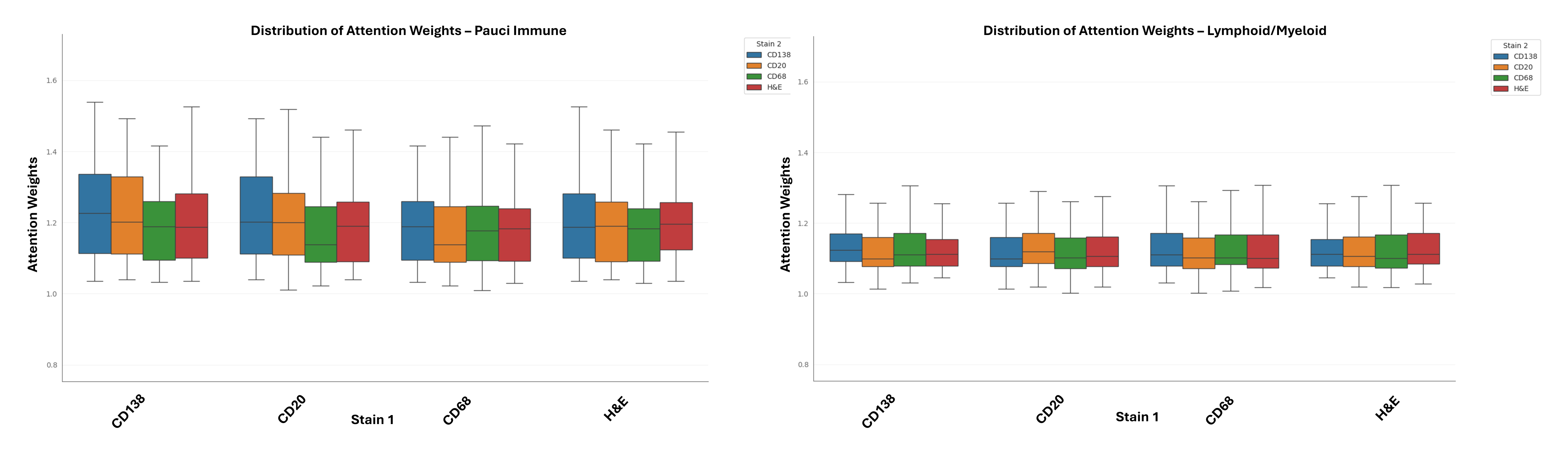}
    \caption{\textbf{Distribution of stain-to-stain interaction} scores for Pauci-Immune (Label 0, left) and Lymphoid/Myeloid (Label 1, right) cases. Each subplot shows the distribution of the average stain-stain attention scores for each stain pair (CD138, CD20, CD68, and H\&E) interact with each other. For each source stain ($x$-axis), the box plots represent the distribution of interaction scores given to each target stain (colored boxes).}
    \label{stain_int_ra}
\end{figure*}

\begin{figure*}[!t]
    \centering
    \includegraphics[width=1\linewidth]{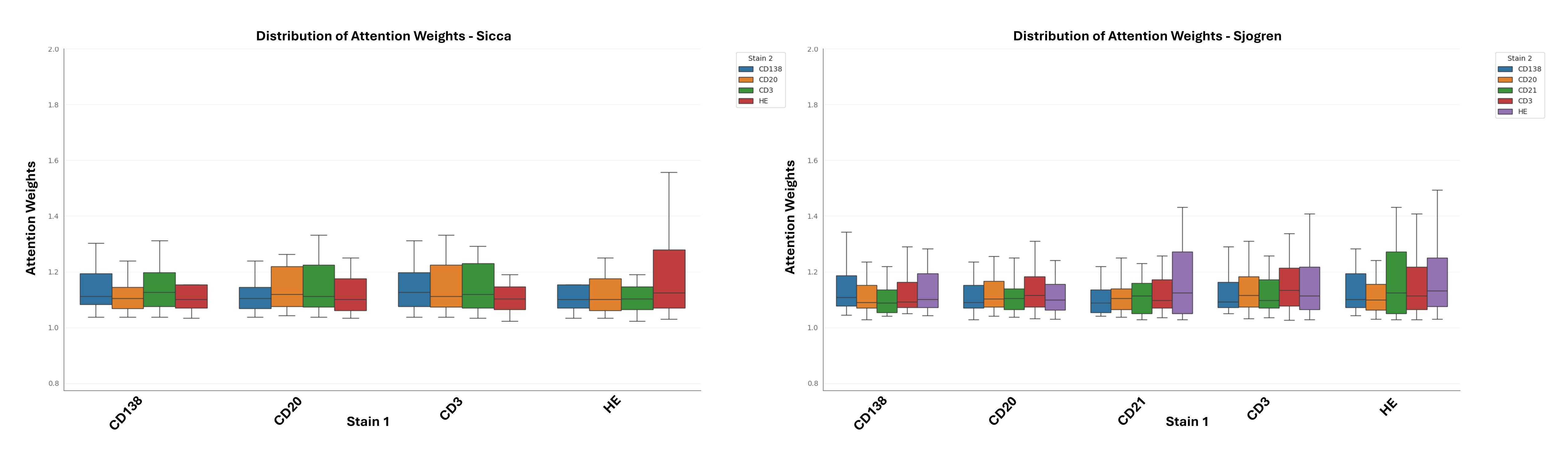}
    \caption{\textbf{Distribution of stain-to-stain interaction} scores for Sicca (Label 0, left) and Sjogren (Label 1, right) cases. Each subplot shows how different stains (CD138, CD20, CD21, CD3, and HE) interact with each other. For each source stain ($x$-axis), the box plots represent the distribution of interaction scores given to each target stain (colored boxes).}
    \label{stain_int_sd}
\end{figure*}

The stain-stain interaction patterns highlight key insights into model decision dynamics, which further deepen our understanding of model behavior and can be linked back to biological mechanisms. These attention-based interactions quantify how the model integrates information across different stain types when making classifications. We present the distribution of stain-stain interactions for both RA and Sjogren's in Figure \ref{stain_int_ra} and \ref{stain_int_sd}.

\subsection{RA}

The stain-stain attention analysis reveals a consistent decrease in all self-interactions (CD138-CD138: $-7.5\%$, CD20-CD20: $-4.7\%$, H\&E-H\&E: $-5.5\%$, CD68-CD68: $-4.5\%$) in Lymphoid/Myeloid compared to Pauci-Immune pathotypes, suggesting a shift from examining intra-stain features toward integrated cross-stain attention patterns, which aligns with the higher entropy scores observed in Lymphoid/Myeloid and the known diffuse inflammatory infiltrates characteristic of this pathotype. The most pronounced changes in cross-stain interactions occur between lymphocyte markers and other stains (CD138-CD20: $-7.4\%$, CD138-H\&E: $-5.3\%$, CD20-H\&E: $-5.3\%$), reflecting the disruption of normal tissue architecture by immune infiltrates in Lymphoid/Myeloid disease. In contrast, macrophage-related interactions (CD68-H\&E: $-4.4\%$, CD138-CD68: $-4.2\%$, CD20-CD68: $-4.0\%$) show more modest changes, suggesting a more consistent role for macrophages across pathotypes. The overall higher and more variable attention weights in Pauci-Immune samples compared to the more uniform, lower weights in Lymphoid/Myeloid indicate that Pauci-Immune classification relies on stronger, more specific feature relationships. Lymphoid/Myeloid requires broader integration of multiple signals, which is consistent with its more complex, heterogeneous inflammatory profile \cite{Lewis2019}.

\subsection{Sjogren}

We see a systematic decrease in self-interactions for Sjogren stain-stain interaction scores (CD20-CD20: $-6.0\%$, CD3-CD3: $-2.2\%$, CD138-CD138: $-1.3\%$), which suggests a shift from paying attention more broadly to the overall context in each single stain, and more toward integrated localized attention spanning across stain types. This aligns with the lower entropy scores obtained for Sjogren stains and the known pathology of more structured lymphoid organization in Sjogren \cite{Kroese2018}. We also note differences in the structural-immune interactions between Sjogren vs Sicca, with differential focus on interactions between HE-IHC stains (increase for HE-CD138 ($+1.9\%$), HE-CD3 ($+1.2\%$); decrease for HE-CD20 ($-4.5\%$)). On the other hand, changes in immune-immune interactions (CD138-CD3: $-2.9\%$, CD138-CD20: $-2.2\%$, CD20-CD3: $-2.2\%$), taken in the context of the balanced stain attention scores obtained for these markers, also suggests a balanced model that integrates information across immune markers.

\section{GNN Heatmaps}
\label{gnn_heatmap}

\begin{figure*}[!t]
    \centering
    \includegraphics[width=1\linewidth]{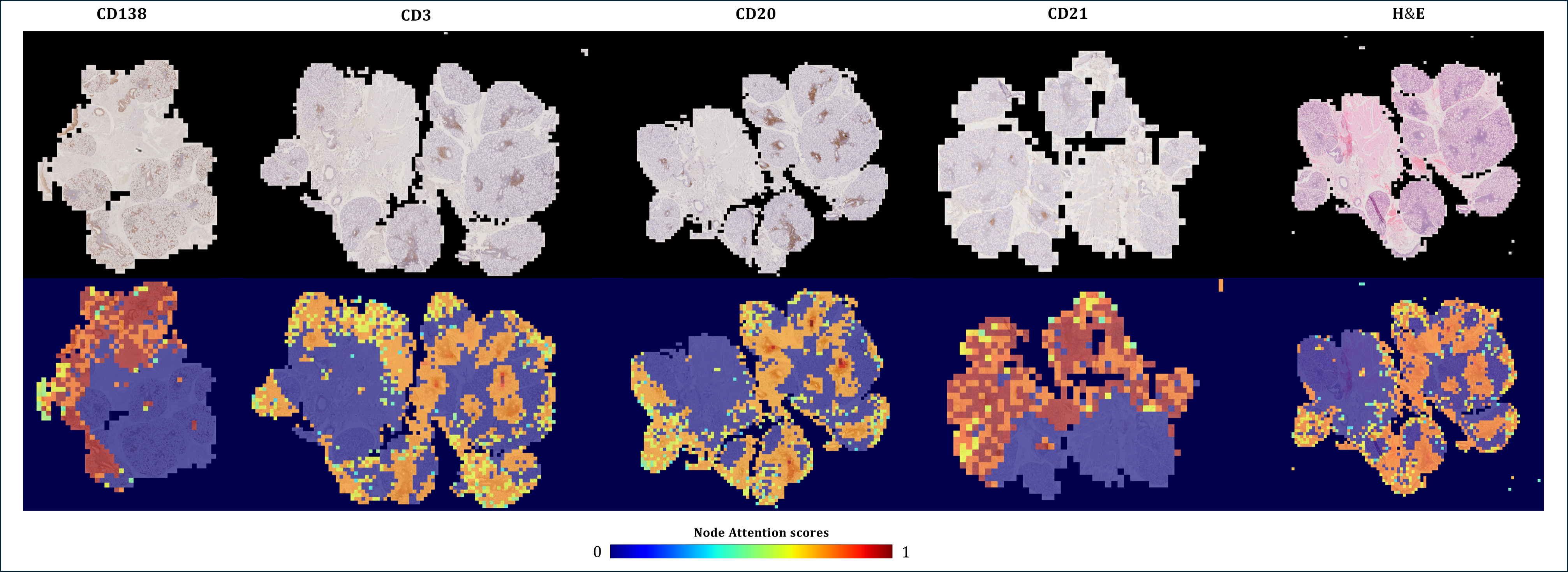}
    \caption{\textbf{Cumulative GNN node attention heatmap} obtained for a Sjogren positive patient with a stack of WSIs consisting of staining for CD138, CD20, C21, CD3 and H\&E. The GNN heatmap corresponds to the direct mapping of the node attention scores back to their original spatial location.}
    \label{heatmap}
\end{figure*}

In Fig. \ref{heatmap}, we show an example of the multistain stack of WSIs (CD138, CD3, CD20, C21, and HE) for one Sjogren positive patient, with the obtained cumulative node attention heatmap for each input stains. The stack of multistain WSIs is the input to our model, and the obtained GNN node heatmaps correspond to the direct mapping of the node attention scores to their original spatial location. We note that our proposed GNN heatmap accurately picks up on the presence of inflammatory aggregates in CD3, CD20, and H\&E, as well as on more disperse attention patterns in CD138 and CD21. CD138 plasma cells are always present throughout the tissue, but will become over-activated and more prevalent in the inflamed tissue, leading to a more diffuse attention pattern. CD21 also accurately focuses on areas with presence of inflammatory aggregates, however also shows a more disperse attention pattern, potentially due to the smaller and fainter aggregates, compared to CD3/CD20 and H\&E. 

\begin{figure*}[!t]
    \centering
    \includegraphics[width=1\linewidth]{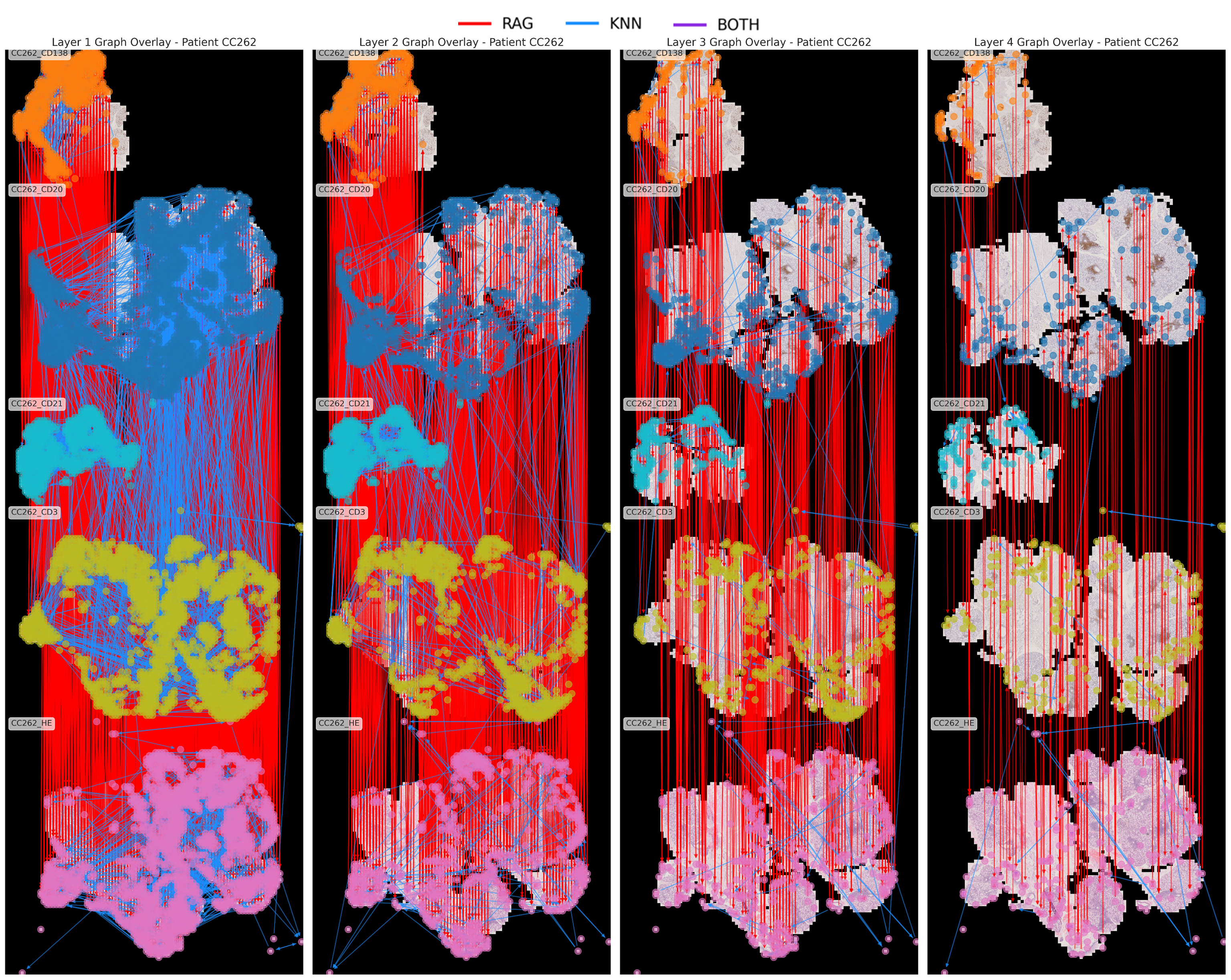}
    \caption{\textbf{Sparsification of input $G_{FRA}$ through the GNN layers.} We plot the multistain patient input graph $G_{FRA}$ as a spatial overlay on the stack of WSIs, to exemplify the connectivity both across and in the WSIs. Edges connect nearest neighbors in both feature (blue) and region adjacent (red) space, with edges which are both feature and region nearest neighbors shown as purple.}
    \label{overlay}
\end{figure*}

To illustrate cross-stack stain-stain interaction and the graph sparsification process through our model, Figure~\ref{overlay} shows $G_{FRA}$ overlaid on the WSIs stack. The layer 1 graph is initially dense with two edge types: region-adjacent edges (red) connecting both across different stains and between spatial neighbors within each WSI, and feature-space edges (blue) linking semantically similar patches regardless of their location. As the graph progresses through the layers, it undergoes progressive sparsification. The transition shows a shift from more homogeneous distributions toward targeted cross-stain interactions, aligning with our quantitative findings of decreased self-attention and enhanced cross-stain integration. By layer 4, the preserved connections highlight important structural-immune relationships between tissue architecture (HE) and immune markers (CD3, CD20, CD21). This progressive refinement demonstrates how the model identifies the organized, integrated nature of immune infiltrates in Sjogren's, capturing diagnostically relevant cross-stain relationships rather than analyzing markers in isolation.

\section{Layer Importance}
\label{layer_importance}

\begin{figure*}[!t]
    \centering
    \includegraphics[width=0.8\linewidth]{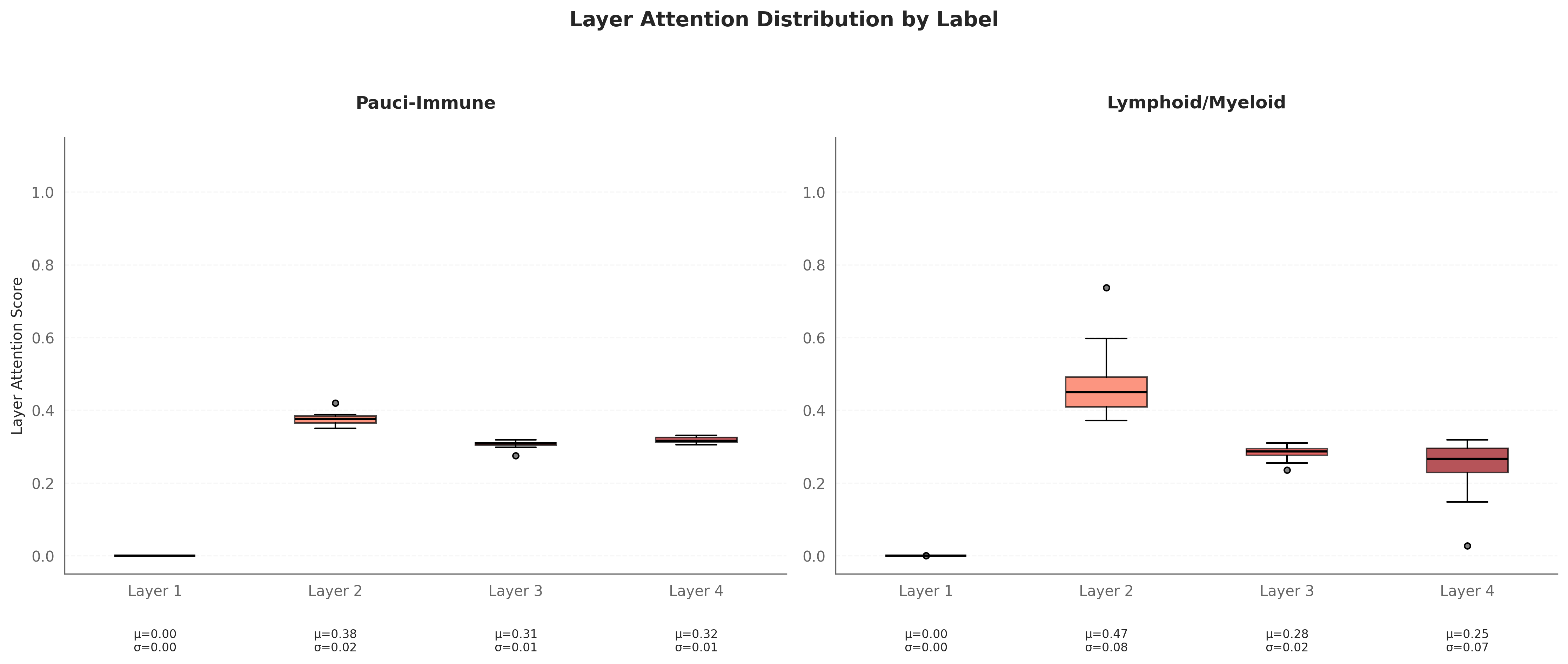}
    \caption{\textbf{Layer-wise attention patterns by label} in the hierarchical graph patient encoder, showing the distribution of attention scores across layers (1-4) for Pauci-Immune and Lymphoid/Myeloid cases, with corresponding mean ($\mu$) and standard deviation ($\sigma$) values.}
    \label{fig:layer_att_ra}
\end{figure*}
\begin{figure*}
    \centering
    \includegraphics[width=1\linewidth]{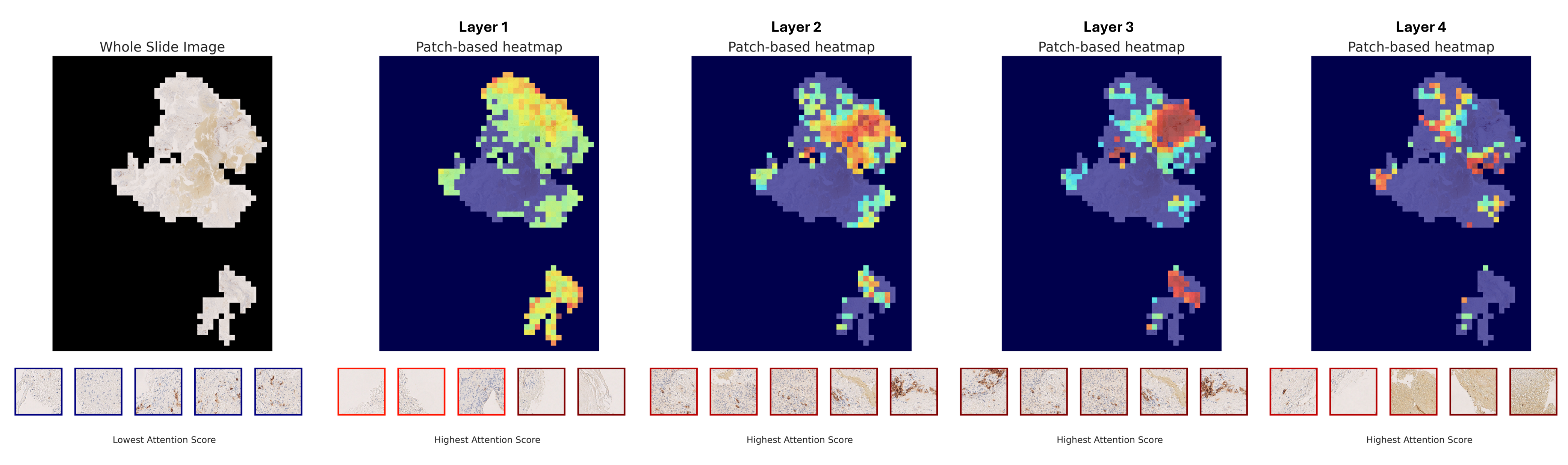}
    \caption{\textbf{Layer-wise attention visualization for a CD138-stained WSI Lymphoid/Myeloid RA patient.} The heatmaps show progression from broad attention in Layer 1 to increasingly focused attention in subsequent layers, with Layer 2 exhibiting the strongest patterns, consistent with quantitative attention scores. Bottom panels show highest and lowest attention patches, revealing cellular infiltrates in Layer 2 and 3.}
    \label{fig:ra_layers}
\end{figure*}
\begin{figure*}[!t]
    \centering
    \includegraphics[width=0.8\linewidth]{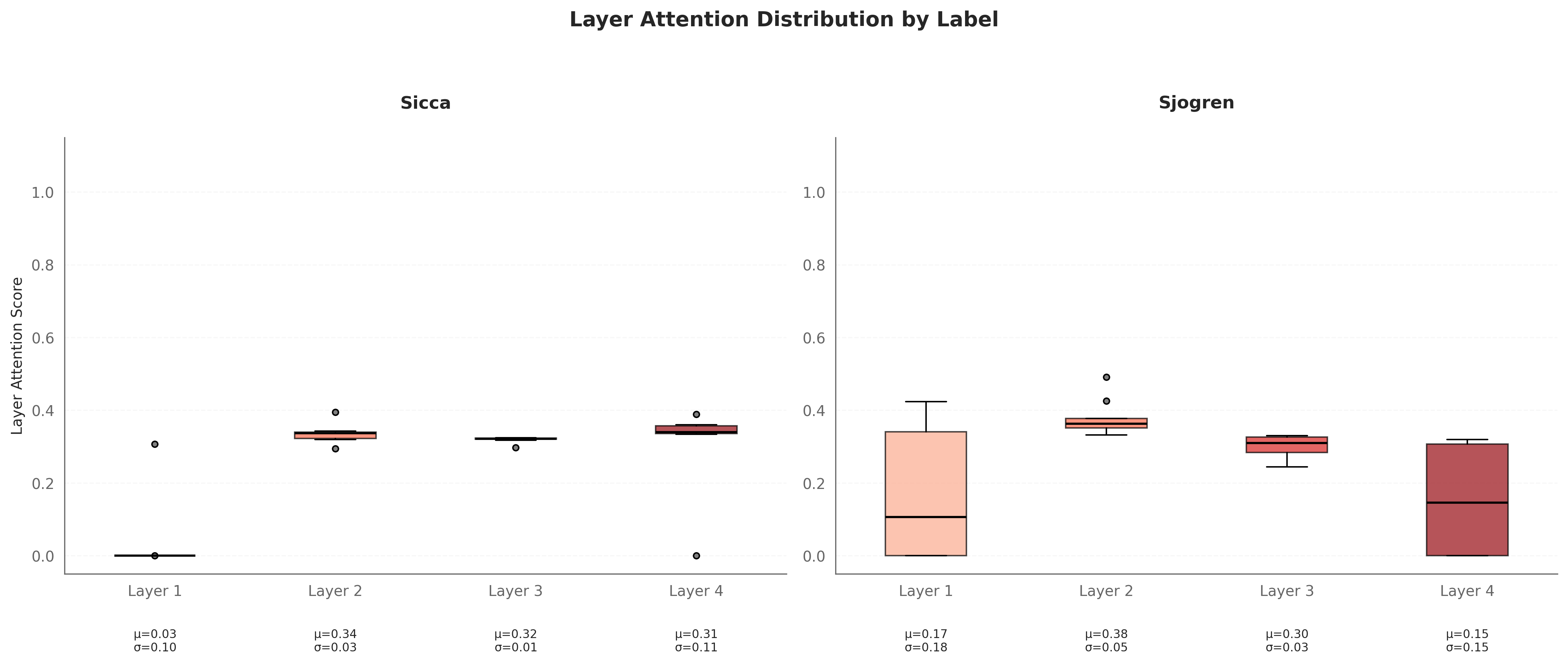}
    \caption{\textbf{Layer-wise attention patterns by label} in the hierarchical graph patient encoder, showing the distribution of attention scores across layers (1-4) for Sicca and Sjogren cases, with corresponding mean ($\mu$) and standard deviation ($\sigma$) values.}
    \label{fig:layer_att_sd}
\end{figure*}

We previously mentioned we chose to maintain a MHSA layer before the classification head in our model architecture, despite seeing a marginal drop in performance. This is because we considered it was a good trade-off with the additional insight  obtained into model decision mechanics, providing another aspect to the explainability of our model with layer importance scores. Briefly, we concatenate the fixed size readouts obtained from each layer of our hierarchical graph patient encoder. This concatenated readout vector is the input to the MHSA. Because we know the size of each layer readout, we can now take the simple step of summing the corresponding attention weights. The rational is this will give us further insight into the role played by each layer in the model decision process and can potentially highlight inherent characteristics on the input data. We present these results in Figures \ref{fig:layer_att_ra} and \ref{fig:layer_att_sd}. 

\subsection{RA}

The layer attention results reveal distinct patterns between pathotypes. Pauci-Immune samples show balanced attention across Layers 2-4 ($\mu=0.38$, $\mu=0.31$, $\mu=0.32$), suggesting reliance on features at multiple abstraction levels. In contrast, Lymphoid/Myeloid samples demonstrate strong preference for Layer 2 ($\mu=0.47$, $\sigma=0.08$), indicating mid-level features are particularly important. This aligns with our stain-stain interaction findings, where Lymphoid/Myeloid showed decreased self-attention and likely depends more on cross-stain integrations occurring at intermediate layers. Both pathotypes assign minimal attention to Layer 1 ($\mu=0.00$), indicating here the raw features have limited classification value without higher-level processing. The higher variance in Layer 2 attention for Lymphoid/Myeloid ($\sigma=0.08$ vs $\sigma=0.02$) suggests greater patient-to-patient variability, consistent with its more heterogeneous inflammatory profile.

To exemplify this process, in Figure \ref{fig:ra_layers} we show the GNN node attention heatmaps obtained for each layer of the model for a WSI with CD138 staining of a RA patient with Lymphoid/Myeloid subtype. We can see a progressive refinement of attention across the layers, with Layer 1 showing broad, diffuse attention across the tissue, while Layers 2-4 reveal increasingly focused attention on specific regions. Layer 2 demonstrates the most pronounced attention patterns, concentrating on areas with visible cellular infiltrates, which aligns with our finding that this layer receives the highest attention weight ($\mu=0.47$) for Lymphoid/Myeloid patients. Layers 3 and 4 further refine this attention, focusing on smaller, more specific regions that likely represent areas with distinctive immune cell aggregates. This visualization supports our quantitative findings and illustrates how the model progressively builds its understanding of the pathotype from general tissue architecture to specific inflammatory aggregates characteristic of Lymphoid/Myeloid disease.

\subsection{Sjogren}

The layer attention distributions reveal distinct hierarchical processing patterns between Sicca and Sjogren's. For Sicca, attention is negligible in Layer 1 ($\mu=0.03$, $\sigma=0.10$) but distributes relatively uniformly across Layers 2-4 ($\mu=0.34$, $\mu=0.32$, $\mu=0.31$ respectively). In contrast, Sjogren's shows substantial Layer 1 attention ($\mu=0.17$, $\sigma=0.18$) followed by peak attention at Layer 2 ($\mu=0.38$, $\sigma=0.05$) and then progressive decline through Layers 3-4 ($\mu=0.30$, $\mu=0.15$), with higher variance observed for Layers 1 and 4. The higher early-layer attention in Sjogren's suggests the model identifies organized immune structures in initial processing stages, corresponding to the decreased self-attention and increased cross-stain integration observed in Sjogren's stain-stain interaction scores. The declining attention pattern in deeper layers for Sjogren's, compared to sustained attention in Sicca, indicates different processing requirements: Sjogren's features are captured earlier through identification of organized lymphoid structures, while Sicca requires more distributed processing across abstraction levels, consistent with its more homogeneous, less structured immune distributions (reflected in higher entropy values).
{\small
\bibliographystyle{ieeenat_fullname}
\bibliography{krag}
}

\end{document}